\documentclass[journal]{IEEEtran}
\usepackage{graphicx}
\usepackage{caption}
\usepackage{amsmath}
\usepackage{algorithm}  
\usepackage{algorithmic} 
\usepackage{multirow}
\usepackage{subfig}
\ifCLASSINFOpdf
\else
\fi
\begin{document}
%
\title{Dual Model Replacement: invisible Multi-target\\Backdoor Attack based on Federal Learning}
%
%
%

\author{Rong Wang†,
        Guichen Zhou,
        Mingjun Gao,
        Yunpeng Xiao,
\thanks{Rong Wang is with School of Software Engineering, Chongqing University of Posts and Telecommunications, Chongqing, 400065, China. E-mail:wangrong1@cqupt.edu.cn}
\thanks{Guichen Zhou is with School of Software Engineering, Chongqing University of Posts and Telecommunications, Chongqing, 400065, China. E-mail:zhouguichen19961030@gmail.com}
\thanks{ Mingjun Gao is with School of Software Engineering, Chongqing University of Posts and Telecommunications, Chongqing, 400065, China. E-mail:gaomj@cqupt.edu.cn}
\thanks{Yunpeng Xiao is with School of Software Engineering, Chongqing University of Posts and Telecommunications, Chongqing, 400065, China. E-mail:xiaoyp@cqupt.edu.cn}
\thanks{† Rong Wang is the corresponding author.}
}

%
%

\markboth{Journal of \LaTeX\ Class Files,~Vol.~1, No.~1, October~2023}%
{Shell \MakeLowercase{\textit{et al.}}: Bare Demo of IEEEtran.cls for IEEE Journals}
%



\maketitle

\begin{abstract}
In recent years, the neural network backdoor hidden in the parameters of the federated learning model has been proved to have great security risks. Considering the characteristics of trigger generation, data poisoning and model training in backdoor attack, this paper designs a backdoor attack method based on federated learning. Firstly, aiming at the concealment of the backdoor trigger, a TrojanGan steganography model with encoder-decoder structure is designed. The model can encode specific attack information as invisible noise and attach it to the image as a backdoor trigger, which improves the concealment and data transformations of the backdoor trigger.Secondly, aiming at the problem of single backdoor trigger mode, an image poisoning attack method called combination trigger attack is proposed. This method realizes multi-backdoor triggering by multiplexing combined triggers and improves the robustness of backdoor attacks. Finally, aiming at the problem that the local training mechanism leads to the decrease of the success rate of backdoor attack, a dual model replacement backdoor attack algorithm based on federated learning is designed. This method can improve the success rate of backdoor attack while maintaining the performance of the federated learning aggregation model. Experiments show that the attack strategy in this paper can not only achieve high backdoor concealment and diversification of trigger forms under federated learning, but also achieve good attack success rate in multi-target attacks.door concealment and diversification of trigger forms but also achieve good results in multi-target attacks.
\end{abstract}

\begin{IEEEkeywords}
Backdoor Attack, Multi-target,Federated Learning,Stealthiness.
\end{IEEEkeywords}

%
\IEEEpeerreviewmaketitle

\section{Introduction}
%
%
%
%
\IEEEPARstart{I}{n} recent years, federated learning \cite{ref1} has been greatly developed as an emerging deep learning framework. To solve the problem of data privacy, the participant's local data and training process cannot be accessed by the server. Because of this feature, malicious attackers can inject the backdoor into the local model by tampering with the local dataset, and then affect the server-side aggregation model by uploading the poisoning model. For researchers, the black box and inexplicability of model parameters also pose great challenges. How to effectively prevent such backdoor attacks is still an urgent problem to be solved. At the same time, the development of attack methods has also attracted great interest from academia and industry.

The backdoor attack is mainly from the following three aspects, respectively, trigger generation, data poisoning, and model training. In the existing trigger generation methods, pixel blocks or translucent patterns \cite{ref2}-\cite{ref5} are mainly added to the image as backdoor triggers. The single fixed feature of the trigger reduces the concealment of the backdoor. Data poisoning is a key step in backdoor attack, which achieves different attack effects by changing the number or style of triggers on the image \cite{ref6}-\cite{ref8}. However, the current backdoor attack schemes often ignore the research on multi-target attacks. With the continuous emergence of new model training methods, there have been studies on backdoor attacks in the model training phase. Federated learning models \cite{ref9}\cite{ref10} have been proven to be vulnerable to backdoor attacks, but these studies did not combine the latest backdoor triggers or data poisoning methods. The backdoor attack scheme for the neural network model under the federated learning training method still has optimization space \cite{ref11}\cite{ref12}, such as the low concealment of the trigger and the single attack target. The goal of the current backdoor attack research is to pursue hidden backdoor triggers, diversified attack targets, and novel backdoor attack model training methods \cite{ref13}-\cite{ref15}.

At present, the trigger has the problem of single attack target. Although the data poisoning method can achieve multi-target attack \cite{ref6}\cite{ref14}, adding multiple visible triggers will greatly reduce the concealment of the backdoor. At the same time, backdoor attacks in federal learning are not efficient. Therefore, the multi-target backdoor attack method based on federated learning still has the following challenges :

Trigger concealment. The amount of hidden data on the trigger has a great influence on the model recognition performance, but improving the concealment of the trigger will reduce the amount of hidden data. How to balance the concealment of triggers and the amount of hidden data is a difficult problem.

Single trigger mode. Under the condition that the total amount of poisoning data is constant, the poisoning data injected into the train set becomes various types, and the segmentation of the poisoning dataset is bound to affect the injection efficiency of the backdoor. How to improve the injection efficiency of the backdoor in the multi-trigger mode becomes a challenge.

Federal learning backdoor attack. The federal learning training strategy is to achieve global model training by aggregating local model parameters. The local backdoor model training process and parameter upload weights determine the effect of backdoor attacks. How to implement the multi-target backdoor injection algorithm of federated learning locally is a problem worthy of study.

Based on the above problems, this paper comprehensively considers three aspects: trigger generation, data poisoning, and model training, and proposes an invisible multi-target backdoor attack method based on federated learning. The contributions of this paper are summarized as follows :

Firstly, the TrojanGan steganography model of encoder-decoder structure is designed. The model combines the image attention mechanism with the DenseNet network structure, hides the attack information coding into the image, and improves the concealment of the backdoor trigger.

Secondly, an image poisoning attack method CTA(Combination Trigger Attack) that can combine triggers is proposed. With the help of the TrojanGan encoder, the attack information is hidden into different half-zones of the image as a single trigger, and the half-zone triggers are combined to form a combined trigger, and a poisoning dataset is formed to create a pre-trained backdoor model.

Finally, the attack algorithm DMR-FLBA(Dual Model Replacement Federal Learning Backdoor Attack) is introduced to ensure that the local backdoor model plays a leading role in the aggregation process. The pre-trained backdoor model is used to replace the federated learning local model, and the backdoor model is modified to upload weights to replace the aggregation model, to maximize the impact of the backdoor model and effectively improve the efficiency of the multi-target backdoor attack of the aggregation model.
\section{RELATED WORK}
Backdoor attack have been an emerging research field in recent years, and DNN is also promoting the development of backdoor attack research. The following article will focus on the three aspects of trigger generation, data poisoning, and federated learning model training, focusing on and analyzing the achievements of researchers in recent years.

First of all, around the steganography of the trigger, to make the backdoor trigger more concealed, the research on the trigger has changed from the visible trigger to the invisible trigger. As a pioneering work of DNN backdoor attack, Gu et al. \cite{ref2} first used visible pixel blocks as backdoor triggers to prove the threat of backdoors to DNN models. Li et al. \cite{ref5} randomly transformed the poisoned image before training and confirmed that the enhanced attack is still available in the physical world. In the subsequent research, Barni et al. \cite{ref3} Liu et al. \cite{ref4} Chen et al. \cite{ref16} tried to embed the trigger into the image in a translucent watermark way, but these triggers are still visible. Li et al. \cite{ref11} realized the steganography of trigger by using an image steganography algorithm and regularization method. Li et al. \cite{ref12} realized the steganography of trigger by using an encoder-decoder structure network. The above article improves the concealment of the trigger through continuous improvement of the trigger generation method. However, there is still room for further optimization between the steganography performance and the steganography capacity.

Secondly, around the data poisoning method, adding triggers to the image through a reasonable method can achieve a more powerful backdoor attack effect. Nguyen et al. \cite{ref17} achieved n-to-one attack effect by generating dynamic pixel triggers based on different images. Xue et al. \cite{ref6} achieved the effect of one-to-n target attack by controlling the pixel value intensity of triggers and achieved the effect of n-to-one attack by combining multiple triggers. Gong et al. \cite{ref18} put the triggers in different positions of the image during model training to eliminate the location restrictions on the triggers during the attack phase. Xiao et al. \cite{ref7} used combined pixels to propose a dynamic trigger using random locations to achieve multi-target attacks. Xue et al. \cite{ref19} achieved the multi-target attack effect by adding a steganography trigger to different channels of the image. The above research has achieved the purpose of multi-target attack by changing the location of triggers or using multiple triggers, but there is still great research value in the way of adding triggers on images.

Finally, focusing on the training method of federated learning model, Bagdasaryan et al. \cite{ref9} and Bhagoji et al. \cite{ref20} first carried out the backdoor attack research under federated learning, confirming that the model after federated learning can be injected into the backdoor, and then Wang et al. \cite{ref21} proposed a marginal backdoor under federal learning. Gong et al. \cite{ref10}, Xie et al. \cite{ref22} proposed a distributed backdoor attack scheme based on the training characteristics of federated learning, which further improved the success rate of backdoor attack. Zhang et al. \cite{ref15}\cite{ref23} proposed the use of generative adversarial networks to generate poisoning data to achieve the purpose of attack by the characteristics of federal learning global shared parameters. Gu et al. \cite{ref38} used the potential representation output of the federated learning participant model to complete the backdoor attack with the help of a small number of labeled samples. The above articles have studied the backdoor attack under federal learning in the poisoning method. At present, the backdoor attack method under federal learning is in the exploratory stage, and the poisoning algorithm still needs to be further explored by researchers.

\section{PROBLEM DEFINITION}
The purpose of this paper is to achieve efficient backdoor attacks by creating high-concealment backdoor triggers, constructing a multi-type poisoning dataset, and implementing a backdoor attack aggregation algorithm under federated learning. First of all, put forward the relevant definition.
\subsection{Related Definition}
\textbf{definition 1} The TrojanGan model training dataset ${{\bf{D}}_{{\rm{invisible}}}} = \{ {{\bf{x}}_i} \in {\bf{X}}|i = 1,...,{\rm{N}}\} $

Training the image dataset required by the TrojanGan network. ${\bf{X}}$ is the sample input space, and $\{ {{\bf{x}}_{\bf{1}}}{\bf{,}}{{\bf{x}}_{\bf{2}}}{\bf{,}}...{\bf{,}}{{\bf{x}}_{\bf{n}}}\} $  is the image sample.

\textbf{definition 2} TrojanGan encoder ${T_{\rm{E}}}$ , decoder ${T_{\rm{D}}}$  and discriminator  ${T_{\rm{C}}}$

 ${T_{\rm{E}}}$ is responsible for receiving the resolution of $W \times H$ the cover image $C$  and the attack information $M \in {\left\{ {0,1} \right\}^{W \times H}}$ as input, and output the steganography image ${C_{{\rm{encode}}}}$ , and then providing the steganography image $image {C_{{\rm{encode}}}}$  to  ${T_{\rm{D}}}$ and ${T_{\rm{C}}}$.${T_{\rm{D}}}$  will output message ${M_{{\rm{decode}}}}$ , if $M = {M_{{\rm{decode}}}}$ , it shows that ${T_{\rm{D}}}$  is valid , at the same time, ${T_{\rm{C}}}$  will receive $C$  and ${C_{{\rm{encode}}}}$  to evaluate the steganography effect and optimize ${T_{\rm{E}}}$  and ${T_{\rm{D}}}$. The relationship between the three is as follows.
 \begin{equation}
     \begin{array}{l}
{T_{\rm{D}}}({C_{{\rm{encode}}}}) = {T_{\rm{D}}}({T_E}(C,M)) = {M_{{\rm{decode}}}}\\
{T_{\rm{C}}}({C_{{\rm{encode}}}}) = {T_{\rm{C}}}({T_E}(C,M))
\end{array}
 \end{equation}

 \textbf{definition 3} image dataset  ${\bf{D}} = \{ ({{\bf{x}}_i},{y_i}) \in {\bf{X}},{\bf{Y}}|i = 1,...,{\rm{N}}\} $

 The image dataset is needed for normal federated learning. The ${\bf{D}}$  is divided proportionally, ${{\bf{D}}_{{\rm{train}}}}$  represents the train set,  ${{\bf{D}}_{{\rm{test}}}}$ represents the test set,  {\bf{X}} represents the sample input space, ${\bf{Y}}$  represents the label space corresponding to the sample, $\{ {{\bf{x}}_{\bf{1}}}{\bf{,}}{{\bf{x}}_{\bf{2}}}{\bf{,}}...{\bf{,}}{{\bf{x}}_{\bf{n}}}\} $  represents the image sample, and $\{ {y_1},{y_2},...,{y_n}\} $  represents the image sample label.

 \textbf{definition 4} Poisoning data subset ${{\bf{D}}_j} = \{ ({{\bf{u}}_i},{v_j}) \in {{\bf{U}}_j},{{\bf{V}}_j}|i \in \{ 1,...,{\rm{N}}\} ,j \in \{ {\rm{s1,s2,cb}}\} \} $

 ${{\bf{D}}_j}$ is a subset of poisoning data that is proportionally extracted from ${{\bf{D}}_{{\rm{train}}}}$  and added triggers and modified labels using ${T_{\rm{E}}}$ . Where $j \in \{ {\rm{s1,s2,cb}}\} $  represents the trigger label, ${{\bf{U}}_j}$  represents the input space of the poisoning sample with the trigger added, ${{\bf{V}}_j} \subset {\bf{Y}}$  represents the label space of the poisoning sample, $\{ {{\bf{u}}_{\bf{1}}}{\bf{,}}{{\bf{u}}_{\bf{2}}}{\bf{,}}...{\bf{,}}{{\bf{u}}_{\bf{n}}}\} $  represents the poisoning image sample, and $\{ {v_{{\rm{s1}}}},{v_{{\rm{s2}}}},{v_{{\rm{cb}}}}\} $  represents the poisoning image sample label. The poisoning dataset ${{\bf{D}}_{\rm{p}}}$  consists of subsets ${{\bf{D}}_{{\rm{s1}}}}$ , ${{\bf{D}}_{{\rm{s2}}}}$ , ${{\bf{D}}_{{\rm{cb}}}}$  and the remaining normal dataset ${{\bf{D}}_{{\rm{remain}}}}$ . Similarly, the  ${{\bf{D}}_{{\rm{p\_test}}}}$ poisoning data test set is the data extracted and processed from ${{\bf{D}}_{{\rm{test}}}}$.
 \begin{equation}
     {{\bf{D}}_{\rm{p}}} = {{\bf{D}}_{{\rm{s1}}}} \cup {{\bf{D}}_{{\rm{s2}}}} \cup {{\bf{D}}_{{\rm{cb}}}} \cup {{\bf{D}}_{{\rm{remain}}}}
 \label{equ2}
 \end{equation}
 \begin{equation}
     {{\bf{D}}_{{\rm{p\_test}}}} = {{\bf{D}}_{{\rm{s1\_test}}}} \cup {{\bf{D}}_{{\rm{s2\_test}}}} \cup {{\bf{D}}_{{\rm{cb\_test}}}}
 \end{equation}
\textbf{ definition 5} Federal learning model  $G_{}^{}$

Let $M \in \{ 1,2...,{\rm{N}}\} $  represent the set of participants. In the $t$  round of model training, the central server will send the current model ${G^t}$  to the selected $m \in M$  participants. The selected local participant $i$  will train the new local model $L_i^{t + 1}$  on the independent local data set ${{\bf{D}}_i}$  and model ${G^t}$ , and then return the updated parameter $L_i^{t + 1} - G_{}^t$  of the local model to the central server for parameter aggregation. The server will average all updates according to its learning rate ${\rm{\eta }}$  to generate a new global model ${G^{t + 1}}$.
\begin{equation}
    G_{}^{t + 1} = G_{}^t + \frac{{\rm{\eta }}}{M}\sum\limits_{i = 1}^{\rm{m}} {(L_i^{t + 1} - G_{}^t)}
\label{equ4}
\end{equation}

\subsection{Problem Formulation}
To describe the research problem of this paper, for the backdoor instance ${\bf{x}}_i^{},i \in \{ 1,2,...,{\rm{N}}\} $  with different attack targets, the attacker tries to maximize the probability of the global model $G_{}^{}$  to predict it as the attack target ${v_j},j \in \{ {\rm{s1,s2,cb}}\} $. $A{C_{train}}$  represents the classification accuracy of the global model on the normal train set ${{\bf{D}}_{{\rm{train}}}}$, $A{C_{\rm{p}}}$  represents the classification accuracy of the backdoor global model on the poisoning train set ${{\bf{D}}_{\rm{p}}}$, and the accuracy fluctuation of the global model cannot exceed the threshold ${\rm{\varphi }}$.Use $R$  to represent the pre-trained backdoor model trained by the poisoning train set ${{\bf{D}}_{\rm{p}}}$, ${X_R}$ to represent the replaced local backdoor model, and the hyperparameter $\beta $ to represent the aggregation weight of the local model $L_{}^{t + 1}$ adjusted by the aggregation algorithm. Finally, the goal of this paper is to obtain the best success rate ${\rm{P}}({G^{t + 1}}({{\bf{D}}_{{\rm{p\_test}}}}))$ of the federated learning aggregation model on ${{\bf{D}}_{{\rm{p\_test}}}}$.
\begin{equation}
    \left. \begin{array}{c}
\max (\sum\limits_{i = 1}^{\rm{N}} {{\rm{P}}[{G^{t + 1}}({\bf{x}}_i^{}) = {v_j}]} )\\
\left| {A{C_{{\rm{train}}}} - A{C_{\rm{p}}}} \right| \le {\rm{\varphi }}\\
L_{}^{t + 1} \approx \beta {\rm{(}}{X_R} - {G^t})
\end{array} \right\} \Rightarrow {\rm{Model}} \Rightarrow {\rm{P}}({G^{t + 1}}({{\bf{D}}_{{\rm{p\_test}}}})) 
\end{equation}
\subsection{Problem Input}
Based on the definition and description in Related Definition , the input of this paper is :

1. TrojanGan model training dataset ${{\bf{D}}_{{\rm{invisible}}}} = \{ {{\bf{x}}_i} \in {\bf{X}}|i = 1,...,{\rm{N}}\} $.

2. Federal learning training dataset ${\bf{D}} = \{ ({{\bf{x}}_i},{y_i}) \in {\bf{X}},{\bf{Y}}|i = 1,...,{\rm{N}}\} $.

3. Poisoning training dataset ${{\bf{D}}_{\rm{p}}} = {{\bf{D}}_{{\rm{s1}}}} \cup {{\bf{D}}_{{\rm{s2}}}} \cup {{\bf{D}}_{{\rm{cb}}}} \cup {{\bf{D}}_{{\rm{remain}}}}$.

4. Poisoning test dataset ${{\bf{D}}_{{\rm{p\_test}}}} = {{\bf{D}}_{{\rm{s1\_test}}}} \cup {{\bf{D}}_{{\rm{s2\_test}}}} \cup {{\bf{D}}_{{\rm{cb\_test}}}}$.

\subsection{Problem Output}
Based on the above description, the following problems are solved by using the model proposed in this paper :

1. TrojanGan encoder ${T_{\rm{E}}}$. The encoder converts the attack information into invisible noise and steals it into the target image as a trigger, and the trigger has good concealment and robustness.

2. Federal learning poisoning dataset ${{\bf{D}}_{\rm{p}}} = {{\bf{D}}_{{\rm{s1}}}} \cup {{\bf{D}}_{{\rm{s2}}}} \cup {{\bf{D}}_{{\rm{cb}}}} \cup {{\bf{D}}_{{\rm{remain}}}}$ and pre-trained backdoor model $R$. By proportionally extracting data from the training dataset  ${{\bf{D}}_{{\rm{train}}}}$ and combining the encoder ${T_{\rm{E}}}$  and CTA data poisoning method to generate ${{\bf{D}}_{{\rm{s1}}}}$,${{\bf{D}}_{{\rm{s2}}}}$ ,${{\bf{D}}_{{\rm{cb}}}}$   as the poisoning sub-dataset, ${{\bf{D}}_{{\rm{remain}}}}$  represents the normal dataset after the extraction, and the pre-trained backdoor model $R$ is trained from the poisoning dataset ${{\bf{D}}_{\rm{p}}}$.

3. Best success rate of backdoor attack ${\rm{P}}({G^{t + 1}}({{\bf{D}}_{{\rm{p\_test}}}}))$. The poisoning test set ${{\bf{D}}_{{\rm{p\_test}}}}$ is used to verify the success rate of the backdoor attack of the federated model. In this paper, the TrojanGan encoder ${T_{\rm{E}}}$ and the test set ${{\bf{D}}_{{\rm{test}}}}$ are used to create the poisoning test set ${{\bf{D}}_{{\rm{p\_test}}}}$, which is input into the trained backdoor aggregation model ${G^{t + 1}}$ to obtain the best backdoor attack success rate.

\section{MODEL}
\begin{figure*}[htbp]
\centering
\includegraphics[scale=0.6]{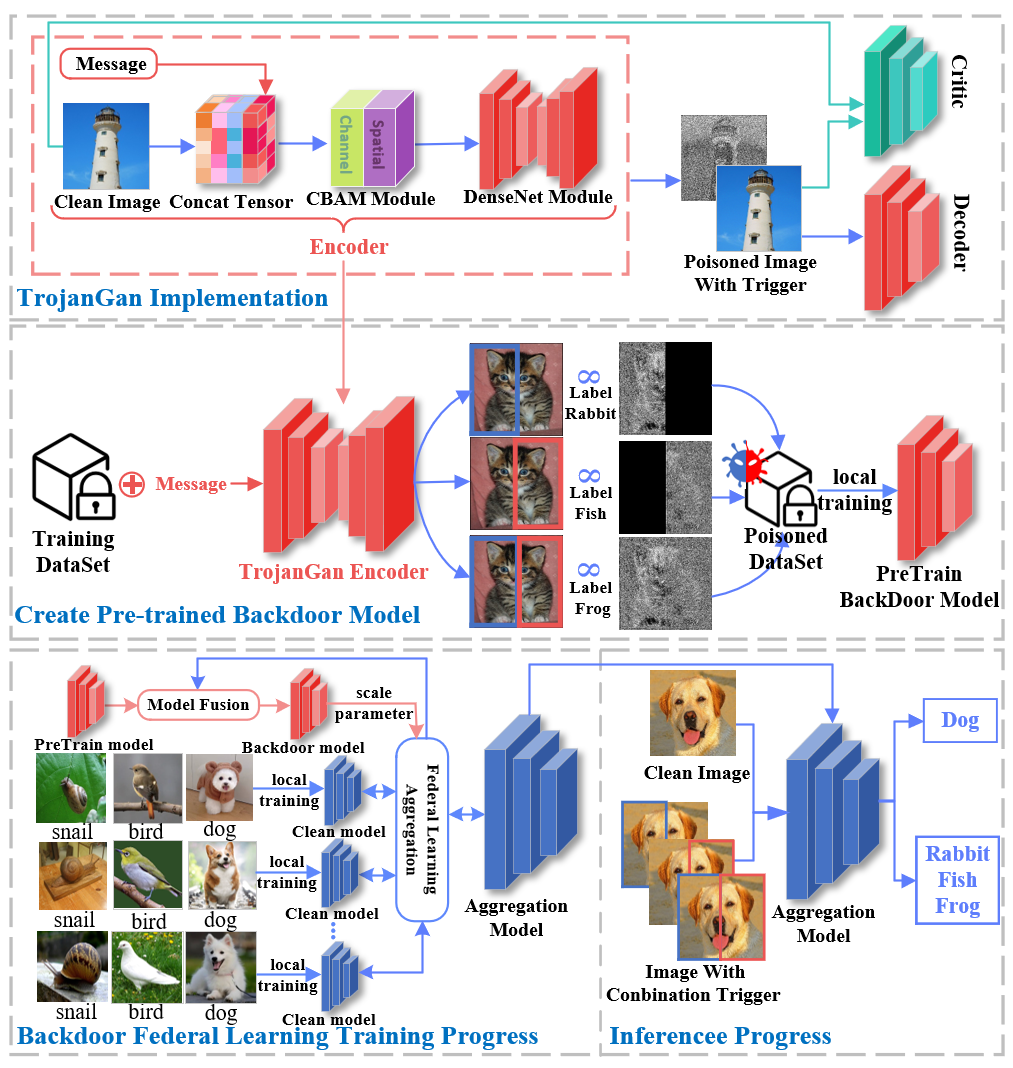}
\caption{\centering{The overall framework of the model}}
\label{fig1}
\end{figure*}
This paper comprehensively considers the attack characteristics of the three stages of trigger generation, data poisoning, and model training in backdoor attacks, and proposes a multi-target backdoor attack method based on federated learning. The overall architecture of the proposed method is shown in Fig.\ref{fig1}. The method is mainly divided into three stages : First, the TrojanGan network is designed to steal the encoded information into the image, generate a steganography trigger, and improve the concealment of the trigger. Then, an image poisoning attack method CTA that can combine triggers is proposed to generate different types of poisoning datasets and create a pre-trained backdoor model. Finally, this paper proposes a dual model replacement backdoor attack algorithm based on federated learning. The dual model replacement method is used to effectively improve the success rate of multi-target backdoor attack.
\subsection{TrojanGan Implementation}
To enhance the concealment of the backdoor trigger, the distribution position and density of the hidden information on the image are optimized. In this paper, a TrojanGan image steganography model based on encoder-decoder structure is constructed. The model includes encoder ${T_{\rm{E}}}$ and decoder ${T_{\rm{D}}}$, which are responsible for optimizing image information encoding and decoding, and discriminator ${T_{\rm{C}}}$, which is responsible for improving image steganography effect.

Firstly, to optimize the distribution of hidden information on the image, the CBAM attention mechanism \cite{ref24} is introduced to construct the Trojan Block module. Secondly, based on the Trojan Block module, the encoder ${T_{\rm{E}}}$, decoder ${T_{\rm{D}}}$ and discriminator ${T_{\rm{C}}}$ of the TrojanGan model are constructed. Finally, the model steganography effect is improved by optimizing multiple training loss functions.
\begin{figure*}[htbp]
\centering
\includegraphics[scale=0.8]{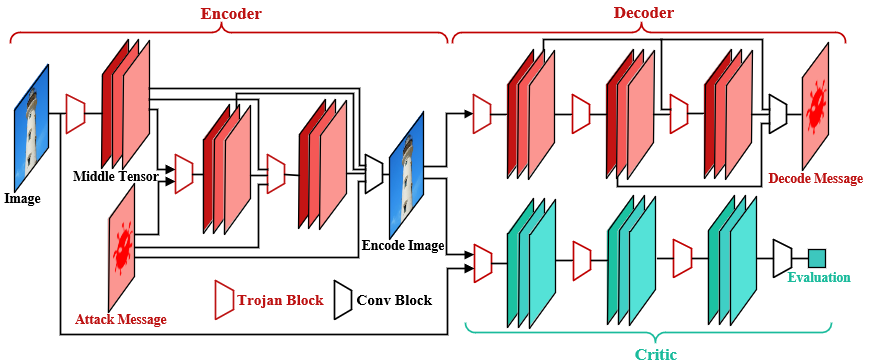}
\caption{\centering{The structure of TrojanGan model}}
\label{fig2}
\end{figure*}

To improve the anti-detection performance of the trigger, this paper adds a CBAM attention layer to the Trojan Block module to focus on the independent features of the image, thereby changing the location and density of hidden information on different images. The Trojan Block module is composed of a convolution kernel of size $3 \times 3$, a convolution layer filled with 1, activation function ReLu, data normalization method BatchNorm2d, and CBAM attention mechanism layer.

To alleviate the problem of gradient disappearance, this paper adopts the network structure of DenseNet \cite{ref25} wind network. The encoder ${T_{\rm{E}}}$ is designed to be composed of 3 Trojan Block modules and 1 output convolution layer, as shown in  Fig.\ref{fig2}. By adding a connection between the encoder modules, the feature mapping generated by the early convolution block is connected to the feature mapping generated by the later convolution block to retain the image features and improve the similarity of the stego image. The encoder is responsible for receiving the $W \times H \times 3$ pixel image and the $W \times H$ data tensor and generating the $W \times H \times 3$ data tensor as a stego image by hiding 1 bit of information in each pixel. The decoder ${T_{\rm{D}}}$ has the same structure as the encoder  ${T_{\rm{E}}}$, and its role is to extract hidden bit information from the image. The decoder receives the $W \times H \times 3$  pixel stego image and decodes the output $W \times H$ data tensor as the hidden data. Since the discriminator only needs to focus on image similarity, the discriminator ${T_{\rm{C}}}$ is designed to be composed of  3 Trojan Block modules and 1 output convolutional layer. The discriminator is used to provide performance feedback to the encoder and decoder, which is responsible for receiving the $W \times H \times 3$  pixel stego image and generating the scalar score of the image.

In this paper, the TrojanGan model is optimized by training iteration, and three loss functions are jointly optimized to optimize the encoder-decoder. To continuously improve the accuracy of encoding-decoding information and calculate the similarity between input information and decoding information, this paper uses cross-entropy loss ${{\cal L}_{\rm{D}}}$ to optimize the decoding accuracy.
\begin{equation}
  {{\cal L}_{\rm{D}}} =  - [M\log {M_{{\rm{decode}}}} + (1 - M)\log (1 - {M_{{\rm{decode}}}})] 
  \label{equ6}
\end{equation}

Image perception loss \cite{ref26} uses deep learning to process the similarity between images, which is more in line with human perception than traditional image similarity calculation methods. To optimize the similarity between the cover image and the stego image, this paper chooses to use LPIPS loss ${{\cal L}_{\rm{P}}}$ and the evaluation model selects vgg.
\begin{equation}
  {{\cal L}_{\rm{P}}} = \frac{1}{{{W_{i,j}} \times {H_{i,j}}}}\sum\limits_{x = 1}^{{W_{i,j}}} {\sum\limits_{y = 1}^{{H_{i,j}}} {{{({\phi _{i,j}}{{(C)}_{x,y}} - {\phi _{i,j}}{{(C{}_{{\rm{encode}}})}_{x,y}})}^2}} } 
  \label{equ7}
\end{equation}
Among them, ${\phi _{i,j}}$ represents the feature map of the pre-training loss model, ${W_{i,j}}$ and ${H_{i,j}}$ represents the dimensions of the feature layer.

To continuously optimize the image generation performance of the encoder and make the generated stego image more realistic, this paper uses the discriminator to judge the authenticity of the image generated by the encoder and uses the discriminator loss ${{\cal L}_{\rm{R}}}$ to optimize the authenticity of the stego image.
\begin{equation}
  {{\cal L}_{\rm{R}}} = {T_{\rm{C}}}({T_{\rm{E}}}(C,M))
\label{equ8}
\end{equation}

The performance of the discriminator determines the authenticity of the encoded image. The Wasserstein loss ${{\cal L}_{\rm{C}}}$  is used to optimize the discriminator, and the score distance of the true and false images is separated to avoid the disappearance of the gradient, allowing the model to continue training.
\begin{equation}
  {{\cal L}_{\rm{C}}} = {T_{\rm{C}}}(C) - {T_{\rm{C}}}({T_{\rm{E}}}(C,M))
\end{equation}

Combined with Formula (\ref{equ6}),(\ref{equ7}),(\ref{equ8}), the joint training loss ${{\cal L}_{\rm{G}}}$ of TrojanGan model is designed. Minimizing the joint training loss is beneficial to improve the coding-decoding accuracy and image steganography performance of the model.${\rm{\lambda }}$ is a hyperparameter.
\begin{equation}
  minimize{{\cal L}_{\rm{G}}} = {\rm{\lambda }}{{\cal L}_{\rm{D}}} + {{\cal L}_{\rm{P}}} + {{\cal L}_{\rm{R}}}
\end{equation}
\subsection{Create Pre-trained Backdoor Model}
To further enhance the harmfulness of backdoor attack and improve the anti-detection of poisoning dataset. In this paper, an image poisoning attack method CTA ( Combination Trigger Attack ) is proposed. By constructing a multi-type poisoning dataset containing combined triggers, different backdoors hidden in the dataset are implanted into the pre-training model, and a multi-target backdoor attack is realized through the reuse of triggers. At the same time, the coincidence characteristics of the combined trigger and the single trigger in different regions alleviate the problem of low backdoor injection efficiency caused by the dispersion of the poisoning dataset.

Firstly, to realize the combined trigger, the position of the image trigger is reset by dividing the image into two different regions. Secondly, independent triggers and combined triggers are set on the two independent regions of the image with the help of the encoder ${T_{\rm{E}}}$ of the TrojanGan model. Finally, the CTA method is proposed to construct the poisoning dataset according to the combination trigger setting method, and the pre-trained backdoor model is created.

For the image sample instance ${\bf{x}} \in {{\bf{D}}_{{\rm{train}}}}$ with resolution $W \times H$, the region of ${\bf{x}}$ in the range of $(0,W/2),(0,H)$ is set to ${{\bf{x}}_{\rm{l}}}$, and the region of  ${\bf{x}}$ in the range of $(W/2,W),(0,H)$ is set to ${{\bf{x}}_{\rm{r}}}$, then ${\bf{x}} = {{\bf{x}}_{\rm{l}}} + {{\bf{x}}_{\rm{r}}}$. ${{\bf{x}}_{\rm{l}}}$ is the left trigger setting area, ${{\bf{x}}_{\rm{r}}}$ is the right trigger setting area, and ${{\bf{x}}_{\rm{l}}} + {{\bf{x}}_{\rm{r}}}$ is the combined trigger setting area.

An equal amount of data samples are randomly selected from each category of $({\bf{x}},y) \in {{\bf{D}}_{{\rm{train}}}}$ to make a poison sub-dataset. Let $\{ {{\rm{t}}_{{\rm{s1}}}},{{\rm{t}}_{{\rm{s2,}}}}{{\rm{t}}_{{\rm{cb}}}}\} $ denote the attack target label and ${\rm{m}}$  denote the attack information. The poisoning sub-dataset ${{\bf{D}}_{s1}}$ is composed of a backdoor image with a left trigger and an attack label. The poisoning sub-dataset ${{\bf{D}}_{s1}}$ is shown in Formula (\ref{equ11}):
\begin{equation}
\begin{split}
     {{\bf{D}}_{{\rm{s1}}}} = \{ ({\bf{x}}_{{\rm{s1}}}^{},{{\rm{t}}_{{\rm{s1}}}})|{\bf{x}}_{{\rm{s1}}}^{} = {T_{\rm{E}}}({{\bf{x}}_{\rm{l}}},m) + {{\bf{x}}_{\rm{r}}},\\
 {{\rm{t}}_{{\rm{s1}}}} \ne y,({\bf{x}},y) \in {{\bf{D}}_{{\rm{train}}}}\} 
\end{split}
\label{equ11}
\end{equation}
Among them, ${\bf{x}}_{{\rm{s1}}}^{} = {T_{\rm{E}}}({{\bf{x}}_{\rm{l}}},m) + {{\bf{x}}_{\rm{r}}}$ represents adding a left trigger on the ${{\bf{x}}_{\rm{l}}}$ region of the original image ${\bf{x}}$,   ${T_{\rm{E}}}$ represents the TrojanGan encoder, $m$ represents the attack information, and $({\bf{x}}_{{\rm{s1}}}^{},{{\rm{t}}_{{\rm{s1}}}})$ represents the generated backdoor image and the attack label.

The poisoning sub-dataset ${{\bf{D}}_{{\rm{s2}}}}$ consists of backdoor image with a right trigger and an attack label. The poisoning sub-dataset ${{\bf{D}}_{{\rm{s2}}}}$ is shown in Formula (\ref{equ12}):
\begin{equation}
\begin{split}
 {{\bf{D}}_{{\rm{s2}}}} = \{ ({\bf{x}}_{{\rm{s2}}}^{},{{\rm{t}}_{{\rm{s2}}}})|{\bf{x}}_{{\rm{s2}}}^{} = {{\bf{x}}_{\rm{l}}} + {T_{\rm{E}}}({{\bf{x}}_{\rm{r}}},m),\\
 {{\rm{t}}_{{\rm{s2}}}} \ne y,({\bf{x}},y) \in {{\bf{D}}_{{\rm{train}}}}\}    
\end{split}
\label{equ12}
\end{equation}
Among them, ${\bf{x}}_{{\rm{s2}}}^{} = {{\bf{x}}_{\rm{l}}} + {T_{\rm{E}}}({{\bf{x}}_{\rm{r}}},m)$ represents adding a left trigger on the ${{\bf{x}}_{\rm{r}}}$ region of the original image ${\bf{x}}$,   ${T_{\rm{E}}}$ represents the TrojanGan encoder, $m$ represents the attack information, and $({\bf{x}}_{{\rm{s2}}}^{},{{\rm{t}}_{{\rm{s2}}}})$ represents the generated backdoor image and the attack label.

The poisoning sub-dataset ${{\bf{D}}_{{\rm{cb}}}}$ is composed of a backdoor image with a combined trigger and an attack tag. The poisoning sub-dataset ${{\bf{D}}_{{\rm{cb}}}}$ is shown in Formula (\ref{equ13}):
\begin{equation}
\begin{split}
 {{\bf{D}}_{{\rm{cb}}}} = \{ ({\bf{x}}_{{\rm{cb}}}^{},{{\rm{t}}_{{\rm{cb}}}})|x_{{\rm{cb}}}^{} = {T_{\rm{E}}}({{\bf{x}}_{\rm{l}}},m) + {T_{\rm{E}}}({{\bf{x}}_{\rm{r}}},m),\\
 {{\rm{t}}_{{\rm{cb}}}} \ne y,({\bf{x}},y) \in {{\bf{D}}_{{\rm{train}}}}\}  
\end{split}
\label{equ13}
\end{equation}
Among them, $x_{{\rm{cb}}}^{} = {T_{\rm{E}}}({{\bf{x}}_{\rm{l}}},m) + {T_{\rm{E}}}({{\bf{x}}_{\rm{r}}},m)$ represents adding a left trigger on the ${{\bf{x}}_{\rm{l}}}$ region of the original image  ${\bf{x}}$, adding a right trigger on the ${{\bf{x}}_{\rm{r}}}$ region, ${T_{\rm{E}}}$  represents the TrojanGan encoder, $m$ represents the attack information, and $({\bf{x}}_{{\rm{cb}}}^{},{{\rm{t}}_{{\rm{cb}}}})$ represents the generated backdoor image and attack label.

The poisoning dataset ${{\bf{D}}_{\rm{p}}}$ is composed of the poisoning sub-datasets ${{\bf{D}}_{{\rm{s1}}}}$, ${{\bf{D}}_{{\rm{s2}}}}$ , ${{\bf{D}}_{{\rm{cb}}}}$  and the remaining normal dataset ${{\bf{D}}_{{\rm{remain}}}}$ where ${{\bf{D}}_{{\rm{reamain}}}} \subset {{\bf{D}}_{{\rm{train}}}}$, namely Formula (\ref{equ2}), $r = {{\left| {{{\bf{D}}_{{\rm{s1}}}} \cup {{\bf{D}}_{{\rm{s2}}}} \cup {{\bf{D}}_{{\rm{cb}}}}} \right|} \mathord{\left/
 {\vphantom {{\left| {{{\bf{D}}_{{\rm{s1}}}} \cup {{\bf{D}}_{{\rm{s2}}}} \cup {{\bf{D}}_{{\rm{cb}}}}} \right|} {\left| {{{\bf{D}}_{{\rm{train}}}}} \right|}}} \right.
 \kern-\nulldelimiterspace} {\left| {{{\bf{D}}_{{\rm{train}}}}} \right|}}$ represent the data poisoning rate.

 To create a pre-trained backdoor model, this paper designs a CTA image poisoning attack method. The basic idea of this method is as follows: First, select the original training dataset ${{\bf{D}}_{{\rm{train}}}}$, and randomly extract the same amount of image samples from each category of ${{\bf{D}}_{{\rm{train}}}}$. Then, the predefined attack information is hidden in these image samples by using the TrojanGan encoder, and the category label of these samples is modified as the target label. Finally, the algorithm repeats the first and second steps three times to construct the poisoning sub-datasets ${{\bf{D}}_{{\rm{s1}}}}$ , ${{\bf{D}}_{{\rm{s2}}}}$ , ${{\bf{D}}_{{\rm{cb}}}}$ , and then combines these three subsets with the remaining normal data set ${{\bf{D}}_{{\rm{remain}}}}$ to form the poisoning dataset ${{\bf{D}}_{\rm{p}}}$ , and uses ${{\bf{D}}_{\rm{p}}}$  to create a pre-trained backdoor model $R$.

 \subsection{Backdoor Federal Learning}
 To improve the injection efficiency and attack success rate of multiple backdoors in the model, while maintaining the normal recognition performance of the federated learning aggregation model. This paper proposes a Dual Model Replacement Federal Learning Backdoor Attack (DMR-FLBA) algorithm. The algorithm uses a dual model replacement method, that is, replacing the local model and modifying the weight of the upload model, to increase the influence of the poisoning parameters in all upload parameters, so that each backdoor still exists in the aggregation model after the average algorithm is diluted.

 Firstly, according to the model training background under federated learning, the ability of the backdoor attacker in this paper is set, that is, the attacker 's control authority over the model training process. Secondly, based on the attacker's ability to design a poisoning terminal dual model replacement backdoor attack method. Finally, a DMR-FLBA federated learning backdoor training algorithm is proposed to improve the efficiency of backdoor attacks under multiple triggers.

 It is set that the attacker has the following control rights under federated learning \cite{ref27}\cite{ref28}: (1) the attacker can completely control the local training dataset ; (2) The attacker can completely control the local training process, such as modifying the local model parameters and updating the training hyperparameters ; (3) Attackers can dynamically adjust the local training process and modify the upload weight of the sub-model. The control rights that the attacker does not have are as follows : (1) The attacker does not have the authority of the federated learning server, that is, it cannot affect the server-side aggregation algorithm ; (2) The attacker cannot influence the local training process of other normal training participants.

\begin{figure*}[htbp]
\centering
\includegraphics[scale=0.5]{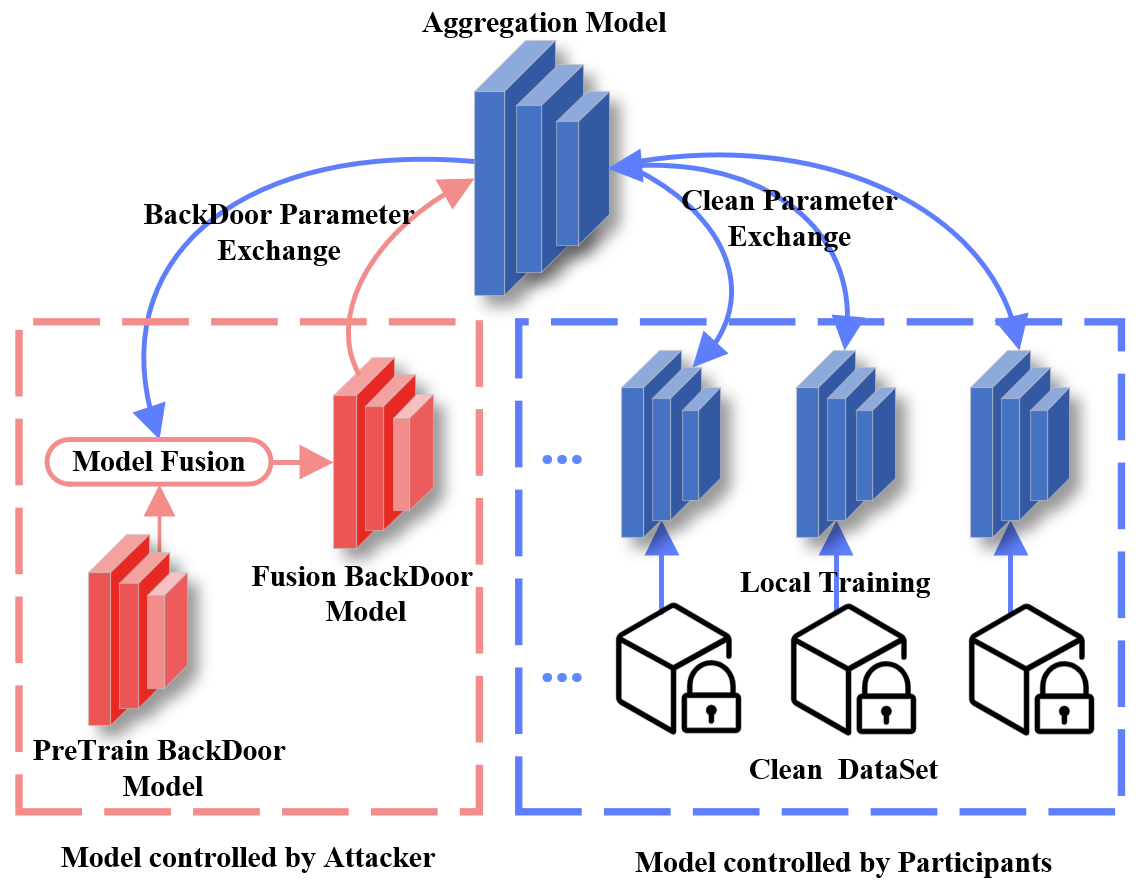}
\caption{\centering{DMR attack}}
\label{fig3}
\end{figure*}

 Fig.\ref{fig3} shows the training process of the backdoor aggregation model under the federated learning in this paper. The red area on the left represents the control authority of the backdoor attacker. The attacker introduces the pre-training model to modify the local model parameters through model fusion, and further modify the uploaded model parameters. The blue area on the right represents normal federated learning participants, who use a predetermined dataset for local training in federated learning.

 Dual model replacement is divided into two steps, backdoor fusion model creation and backdoor fusion model parameter upload. First, the attacker does not perform local model training and fuses the pre-trained backdoor model with the aggregation model to form a backdoor fusion model for parameter upload, ensuring the recognition performance of the local backdoor model on normal dataset and different types of poisoning datasets, and ensuring that the local backdoor model does not deviate too much from the aggregation model. The model fusion method is:
 \begin{equation}
{X_R} = R + \alpha {G^t}
\label{equ14}
\end{equation}
Among them, ${X_R}$ represents the backdoor fusion model, $R$ represents the backdoor pre-training model, ${G^t}$ represents the aggregation model, and $\alpha $ represents the hyperparameter.

For a backdoor fusion model ${X_R}$ in federated learning, as shown in Formula (\ref{equ14}), the model injects multiple backdoors locally at the same time through the first model replacement and then performs the second model replacement. The aggregation model ${G^{t + 1}}$ in Formula (\ref{equ4}) is replaced by the backdoor fusion model ${X_R}$ to obtain
\begin{equation}
{X_R} = G_{}^t + \frac{{\rm{\eta }}}{M}\sum\limits_{i = 1}^{\rm{m}} {(L_i^{t + 1} - G_{}^t)} 
\label{equ15}
\end{equation}
Among them, for the normal local model $L_i^{t + 1},i = 1,...,{\rm{m}} - 1$, when the model is close to convergence, we obtain
\begin{equation}
\sum\limits_{i = 1}^{{\rm{m - 1}}} {(L_i^{t + 1} - G_{}^t)}  \approx 0
\label{equ16}
\end{equation}
comprehensive formula (\ref{equ15}) and (\ref{equ16}) get the formula,
\begin{equation}
L_{\rm{m}}^{t + 1} \approx \frac{M}{{\rm{\eta }}}({X_R} - {G^t}) + {G^t}
\label{equ17}
\end{equation}
Among them, by scaling the hyperparameter $\beta {\rm{ = }}{M \mathord{\left/
 {\vphantom {M {\rm{\eta }}}} \right.
 \kern-\nulldelimiterspace} {\rm{\eta }}}$, the upload weight of the parameters of the backdoor fusion model is affected. In the actual attack process, the hyperparameter $\beta $ will be much larger than 1. To further expand the influence of the backdoor fusion model, the formula (\ref{equ17}) can be further simplified to:
 \begin{equation}
L_{\rm{m}}^{t + 1} \approx \beta ({X_R} - {G^t})
\label{equ18}
\end{equation}
When the poisoning terminal uploads model $L_{\rm{m}}^{t + 1}$, the influence of the poisoning model in the aggregation model can be increased.

To implant multiple backdoors into the aggregation model, this paper designs a dual model replacement backdoor attack algorithm DMR-FLBA based on federated learning. The basic idea of the algorithm is as follows: Firstly, the backdoor pre-training model generated according to the CTA method is loaded. Secondly, the first model replacement is performed. According to Formula (\ref{equ14}), the backdoor pre-training model is combined with the global aggregation model to form a backdoor fusion model. Finally, the second model replacement is carried out, and the upload weight of the local backdoor fusion model parameters is adjusted according to Formula (\ref{equ18}), to enhance the influence of the poisoning parameters in the aggregation model and improve the success rate of multi-target backdoor attack.

 \subsection{Algorithm Implementation}
In this section, we introduce the dual model replacement backdoor attack algorithm DMR-FLBA based on federated learning. The algorithm includes using the TrojanGan encoder to add multiple types of backdoor triggers and proportionally build a poisoning dataset, creating a pre-trained backdoor model, using the pre-trained backdoor model to create a backdoor fusion model, and adjusting the parameter upload weight. The DMR-FLBA algorithm is shown in the following Algorithm 1:

\begin{algorithm}[htb]
	\caption{    DMR-FLBA.}
	\label{xxx}
	\begin{algorithmic}[1] 
		\REQUIRE ~~\\ 
		Training dataset ${{\bf{D}}_{{\rm{train}}}}$, trojanGan encoder ${T_{\rm{E}}}$,
  
        aggregation model $G_{}^t$;
		\ENSURE ~~\\ 
		backdoor aggregation model;
		\STATE Import the TrojanGan encoder ${T_{\rm{E}}}$
		\STATE Create a backdoor image dataset ${{\bf{D}}_{{\rm{s1}}}}$ for left trigger based on Formula (11) 
        \STATE Create a backdoor image dataset ${{\bf{D}}_{{\rm{s2}}}}$  for right trigger based on Formula (12) 
		\STATE Create a backdoor image dataset ${{\bf{D}}_{{\rm{cb}}}}$  for combined trigger based on Formula (13) 
        \STATE Combine data to form a poisoning dataset ${{\bf{D}}_{\rm{p}}}$  based on Formula (2)
        \STATE reate a backdoor pre-training model $R$  based on ${{\bf{D}}_{\rm{p}}}$ 
        \STATE Replace the local model with the pre-trained model $R$
        \STATE Create a backdoor fusion model based on Formula (14)
        \STATE Adjust the parameters of the uploaded backdoor fusion model based on Formula (18) 
        \STATE The federated learning aggregation model performs parameter fusion
	\end{algorithmic}
\end{algorithm}

In the above algorithm, the number of local iterative training of normal participants is $E$ and the number of global iterative training of federated learning is $N$. When the algorithm is executed, it is necessary to extract data from the train set ${{\bf{D}}_{{\rm{train}}}}$, and the time complexity of creating the poisoning dataset is $O(N)$. The time complexity of federated learning training is $O(E \times N)$, so the time complexity of the whole model training is $O(E \times N) + O(N)\sim O(E \times N)$.

\section{EXPERIMENT}
\subsection{Experiment Setting}
\subsubsection{Datasets and Models}
In this experiment, the effectiveness of federated learning on three different datasets was evaluated. The image classification public dataset CIFAR10, GTSRB, and the large real face dataset MS-Celeb-1M were used respectively. ResNet-18 \cite{ref29} was used as the model structure of these three datasets.

\textbf{CIFAR10} \cite{ref30}: The original data set contains 60,000 images. Image classification includes 10 categories of animals and vehicles, of which 50,000 images are used for training ( 5000 images for each category ) and 10,000 images are used for testing ( 1000 images for each category ). The image size is 3 × 32 × 32.

\textbf{GTSRB} \cite{ref31}: German traffic sign recognition image data, the original data contains about 50,000 images, 43 categories. Among them, 39,252 images are used for training ( 200-2000 images per class ), and about 12,630 images are used for testing. The image size is adjusted to 3 × 32 × 32.

\textbf{MS-Celeb-1M} \cite{ref32}: In the original dataset, there are nearly 100,000 identities, each of which contains about 100 images. Because the image identity in the original data set is chaotic, the method provided in the article \cite{ref33} is used to clean the original dataset. A total of 26,000 images of 200 identities were selected from the cleaned dataset, of which 24,000 images were used for training ( 120 images per class ) and 2,000 images for testing ( 10 images per class ). The image size was adjusted to 3 × 64 × 64.
\subsubsection{Baseline Selection}
This paper compares the proposed backdoor attack method with the existing backdoor attack method. Five baseline methods are described as follows :

\textbf{BadNet} \cite{ref2} first proposed the use of visible pixel blocks as triggers.

\textbf{WaNet} \cite{ref11.5} used an invisible trigger based on image distortion and proposed a training mode called ‘noise’.

\textbf{Poison Ink} \cite{ref13} implemented a sample-specific backdoor trigger by extracting image edges using neural network steganography.

\textbf{Distributed Backdoor Attack} \cite{ref22} proposed a distributed backdoor attack method in federated learning for the first time.

\textbf{Coordinated Backdoor Attacks} \cite{ref10} enable attackers in federated learning to generate model-dependent local triggers for distributed backdoor attack based on the aggregated model received from the server.

\textbf{LR-BA} \cite{ref38} uses the potential representation output of the federated learning participant model to complete the backdoor attack with the help of a small number of labeled samples.
\subsubsection{Evaluation Metric}
To evaluate the effectiveness of the backdoor attack, this paper will use the following evaluation methods.

\textbf{Image steganography performance}. To evaluate the similarity between normal images and stego images, this paper uses peak signal-to-noise ratio (PSNR) \cite{ref34}, structural similarity (SSIM) \cite{ref35} and ${\ell ^\infty }$ norm \cite{ref36} as the evaluation criteria of Steganography performance.
 \begin{equation}
PSNR = 10\log \frac{{{{({2^n} - 1)}^2}}}{{MSE}}
\label{equ19}
\end{equation}
 \begin{equation}
SSIM = \frac{{(2{\mu _X}{\mu _Y} + {k_1}R)(2{\sigma _{XY}} + {k_2}R)}}{{(\mu _X^2 + \mu _Y^2 + {k_1}R)(\sigma _X^2 + \sigma _Y^2 + {k_2}R)}}
\label{equ20}
\end{equation}
 \begin{equation}
{\ell ^\infty } = {\max _{1 \le i \le n}}\sum\nolimits_{j = 1}^n {|{a_{ij}}|} 
\label{equ21}
\end{equation}

\textbf{Attack Success Rate (ASR)}. is defined as the accuracy of the backdoor model to test the backdoor instance. Effective backdoor attack strategy can improve the success rate of backdoor attack

\textbf{Benign Accuracy ( BA )}. is defined as the accuracy of the backdoor model to test normal instances. The higher the BA, the better the classification effect of the backdoor model on normal instances.

\begin{table*}[htbp]
\caption{\centering{The structure of TrojanGan model}}
\begin{tabular}{c|l|c|c|c|c|c|c|c}
\hline
Module Name              & \multicolumn{1}{c|}{Layer Name}    & Kernel & Padding & Channel I/O & Normalization & Activation & Attention & Input                                                                                \\ \hline
\multirow{4}{*}{Encoder} & \multicolumn{1}{c|}{Trojan Block1} & 3×3    & 1       & 3/32        & BatchNorm2d   & ReLU       & CBAM      & Images                                                                               \\ \cline{2-9} 
                         & Trojan Block2                      & 3×3    & 1       & 33/32       & BatchNorm2d   & ReLU       & CBAM      & Trojan Block1+Data                                                                   \\ \cline{2-9} 
                         & Trojan Block3                      & 3×3    & 1       & 65/32       & BatchNorm2d   & ReLU       & CBAM      & Trojan Block1+Trojan Block2                                                          \\ \cline{2-9} 
                         & \multicolumn{1}{c|}{Conv4}         & 3×3    & 1       & 97/3        & -             & -          & -         & \begin{tabular}[c]{@{}c@{}}Trojan Block1+Trojan Block2+\\ Trojan Block3\end{tabular} \\ \hline
\multirow{4}{*}{Decoder} & Trojan Block5                      & 3×3    & 1       & 3/32        & BatchNorm2d   & ReLU       & CBAM      & Conv4                                                                                \\ \cline{2-9} 
                         & Trojan Block6                      & 3×3    & 1       & 32/32       & BatchNorm2d   & ReLU       & CBAM      & Trojan Block5                                                                        \\ \cline{2-9} 
                         & Trojan Block7                      & 3×3    & 1       & 64/32       & BatchNorm2d   & ReLU       & CBAM      & Trojan Block5+Trojan Block6                                                          \\ \cline{2-9} 
                         & \multicolumn{1}{c|}{Conv8}         & 3×3    & 1       & 96/1        & -             & -          & -         & \begin{tabular}[c]{@{}c@{}}Trojan Block5+Trojan Block6+\\ Trojan Block7\end{tabular} \\ \hline
\multirow{4}{*}{Critic}  & Trojan Block9                      & 3×3    & -       & 3/32        & BatchNorm2d   & ReLU       & CBAM      & Images                                                                               \\ \cline{2-9} 
                         & Trojan Block10                     & 3×3    & -       & 32/32       & BatchNorm2d   & ReLU       & CBAM      & Trojan Block9                                                                        \\ \cline{2-9} 
                         & Trojan Block11                     & 3×3    & -       & 32/32       & BatchNorm2d   & ReLU       & CBAM      & Trojan Block10                                                                       \\ \cline{2-9} 
                         & \multicolumn{1}{c|}{Conv12}        & 3×3    & -       & 32/1        & -             & -          & -         & Trojan Block11                                                                       \\ \hline
\end{tabular}
\label{table1}
\end{table*}

\begin{table*}[ht]
\caption{\centering{Image steganography performance}}
\centering
\begin{tabular}{|c|ccc|lll|lll|}
\hline
DataSets   & \multicolumn{3}{c|}{CIFAR10}                                    & \multicolumn{3}{c|}{GTSRB}                                                        & \multicolumn{3}{c|}{MS-Celeb-1M}                                                  \\ \hline
Model      & \multicolumn{1}{c|}{PSNR↑} & \multicolumn{1}{c|}{SSIM↑} & ${\ell ^\infty }$↓    & \multicolumn{1}{c|}{PSNR↑} & \multicolumn{1}{c|}{SSIM↑} & \multicolumn{1}{c|}{${\ell ^\infty }$↓} & \multicolumn{1}{c|}{PSNR↑} & \multicolumn{1}{c|}{SSIM↑} & \multicolumn{1}{c|}{${\ell ^\infty }$↓} \\ \hline
BadNet     & \multicolumn{1}{c|}{26.43} & \multicolumn{1}{c|}{0.926} & 186.2 & \multicolumn{1}{l|}{25.91} & \multicolumn{1}{l|}{0.899} & 199.6                   & \multicolumn{1}{l|}{24.54} & \multicolumn{1}{l|}{0.948} & 203                     \\ \hline
WaNet     & \multicolumn{1}{c|}{21.19} & \multicolumn{1}{c|}{0.897} & 141.2 & \multicolumn{1}{l|}{23.93} & \multicolumn{1}{l|}{0.89} & 129                   & \multicolumn{1}{l|}{24.16} & \multicolumn{1}{l|}{0.902} & 146.7                     \\ \hline
Poison Ink & \multicolumn{1}{c|}{42.95} & \multicolumn{1}{c|}{0.996} & 14.35 & \multicolumn{1}{l|}{41.76} & \multicolumn{1}{l|}{0.995} & 16.50                   & \multicolumn{1}{l|}{42.68} & \multicolumn{1}{l|}{0.994} & 18.14                   \\ \hline
TJG-LT     & \multicolumn{1}{c|}{37.56} & \multicolumn{1}{c|}{0.986} & 21.16 & \multicolumn{1}{l|}{36.56} & \multicolumn{1}{l|}{0.972} & 23.45                   & \multicolumn{1}{l|}{40.39} & \multicolumn{1}{l|}{0.986} & 17.05                   \\ \hline
TJG-RT    & \multicolumn{1}{c|}{37.48} & \multicolumn{1}{c|}{0.985} & 21.39 & \multicolumn{1}{l|}{36.30} & \multicolumn{1}{l|}{0.973} & 24.03                   & \multicolumn{1}{l|}{40.28} & \multicolumn{1}{l|}{0.987} & 17.18                   \\ \hline
TJG-CT    & \multicolumn{1}{c|}{35.04} & \multicolumn{1}{c|}{0.976} & 21.96 & \multicolumn{1}{l|}{33.80} & \multicolumn{1}{l|}{0.952} & 24.96                   & \multicolumn{1}{l|}{37.74} & \multicolumn{1}{l|}{0.977} & 17.69                   \\ \hline
\end{tabular}
\label{table2}
\end{table*}

\begin{figure*}[htbp]
\centering
\includegraphics[scale=0.7]{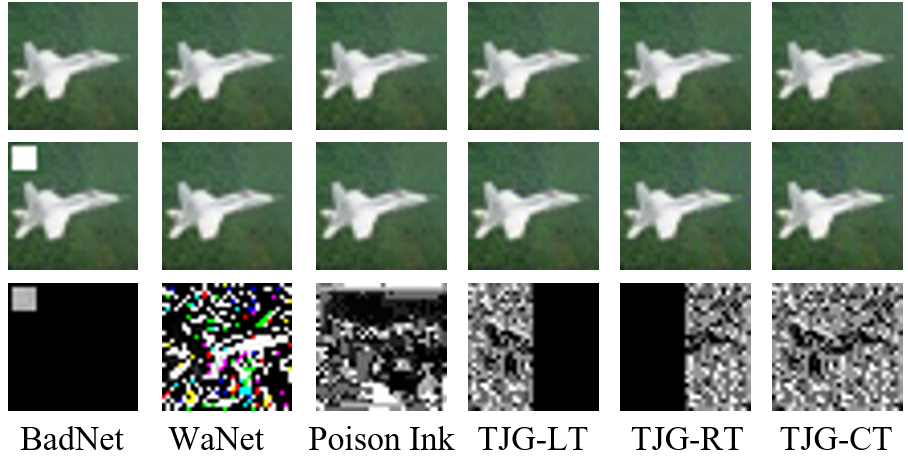}
\caption{\centering{Comparison of triggers}}
\label{fig4}
\end{figure*}

\subsection{Experimental Analysis}
In this section, we first show the structure and training parameters of the TrojanGan model and compare the steganography performance of the baseline method. Then, the TrojanGan model is used to create a poisoning dataset, and the performance of the pre-trained model is created and analyzed. Finally, the backdoor attack effect of the DMR-FLBA algorithm in federated learning under different parameters is analyzed.

\subsubsection{Steganography Performance Analysis}
According to Section TrojanGan Implementation, training is performed using the structure given in Table \ref{table1}. The TrojanGan model train set ${{\bf{D}}_{{\rm{invisible}}}}$ consists of 10,000 ImageNet \cite{ref37} train set images, and the validation set consists of 1,000 ImageNet test set images. The TrojanGan model uses an Adam optimizer with a learning rate of $1{e^{ - 4}}$, hyperparameters ${\rm{\lambda  = 10}}$, training epoch = 16.

For the baseline model, the Poison Ink injection network and the guidance extraction network are trained according to the article \cite{ref13}]. The train set consists of 25,000 CIFAR10 train set images, and the validation set consists of 5000 CIFAR10 test set images. The training epoch = 200.

Fig.\ref{fig4} shows the comparison of different triggers on the CIFAR10 image. The first line is the original image, the second line is the stego image with the trigger, and the third line is the residual image. TJG-LT, TJG-RT, and TJG-CT are used to represent the TrojanGan left trigger, right trigger, and combined trigger identification, respectively. By analyzing the residual image, in addition to using the visible pixel trigger of BadNet, WaNet creates a steganography trigger by distorting the edge of the object through random noise, Poison Ink creates a steganography trigger by focusing on the contour edge of the object, and TrojanGan creates a steganography backdoor trigger by focusing on the object itself. By observing the steganography image in the second row of Fig.\ref{fig4}, it can be concluded that the steganography image generated by the TrojanGan model also achieves the same effect that the human eye cannot distinguish compared with other steganography methods.

Table \ref{table2} shows the image steganography performance, and each value is composed of the average value of 1000 stego images compared with the original image.The ${\ell ^\infty }$ norm of WaNet is smaller than that of BadNet, but PSNR and SSIM are similar to BadNet, showing greater damage to the image structure. Both Poison Ink and TrojanGan improve the steganography ability through neural network steganography. The steganography performance of TJG-LT and TJG-RT single-side trigger is better than that of TJG-CT combined trigger. However, compared with Poison Ink, TrojanGan has an average difference of 4 on PSNR, 0.01 on SSIM, and 4.47 on ${\ell ^\infty }$ norm. Compared with Poison Ink, the TrojanGan model improves the amount of steganography data to ensure the recognition efficiency of the backdoor, but sacrifices some steganography performance.

\subsubsection{Pre-trained Model Performance Analysis}
The default setting of pre-trained model attack : ResNet18 is used as the model structure, and the poisoning dataset is constructed before the model training. The total poisoning rate of CIFAR10 and GSTRB is 10\%, and the total poisoning rate of MS-Celeb-1M is 15\%. Among them, the three types of poisoning sub-datasets account for 3.33\% and 5\%, respectively. The poisoning test set is half of the normal test set (3000-5000), and the multi-type attack labels are 0,1,2, respectively. The TrojanGan steganography information is ‘goldfish’. The pre-trained backdoor model has a total training epoch = 80. The learning rate is lr = 0.01 in the first 50 rounds and then adjusted to 0.001. The optimizer selects SGD, momentum = 0.9, weight\_decay = 5e-4. After each round of training, the model with the best performance on the poisoning test set is selected to save.

For the training baseline model, WaNet performs model training through noise mode, using ResNet18 as the model structure, epoch = 600 on CIFAR10, epoch = 300 on GTSRB and MS-Celeb-1M. Other baseline model training settings are the same as the pre-training model, but the attack label is set to 0..

\begin{table}[ht]
\caption{\centering{Performance comparison of pre-trained backdoor models}}
\centering
\scalebox{0.9}{
\begin{tabular}{|c|cc|cc|cc|}
\hline
DataSets    & \multicolumn{2}{c|}{CIFAR10}                          & \multicolumn{2}{c|}{GTSRB}                            & \multicolumn{2}{c|}{MS-Celeb-1M}                      \\ \hline
Model       & \multicolumn{1}{c|}{BA(\%)}                 & ASR(\%) & \multicolumn{1}{c|}{BA(\%)}                 & ASR(\%) & \multicolumn{1}{c|}{BA(\%)}                 & ASR(\%) \\ \hline
Clean Model & \multicolumn{1}{c|}{85.42}                  & -       & \multicolumn{1}{c|}{94.80}                  & -       & \multicolumn{1}{c|}{91.10}                  & -       \\ \hline
BadNet      & \multicolumn{1}{c|}{83.93}                  & 100.0   & \multicolumn{1}{c|}{94.90}                  & 100.0   & \multicolumn{1}{c|}{90.30}                  & 100.0   \\ \hline
WaNet      & \multicolumn{1}{c|}{86.70}                  & 90.98   & \multicolumn{1}{c|}{96.2}                  & 98.59   & \multicolumn{1}{c|}{91.75}                  & 99.25   \\ \hline
Poison Ink  & \multicolumn{1}{c|}{84.03}                  & 99.86   & \multicolumn{1}{c|}{94.35}                  & 99.83   & \multicolumn{1}{c|}{89.64}                  & 98.55   \\ \hline
TJG-LT      & \multicolumn{1}{c|}{\multirow{3}{*}{84.44}} & 100.0   & \multicolumn{1}{c|}{\multirow{3}{*}{94.94}} & 99.90   & \multicolumn{1}{c|}{\multirow{3}{*}{89.35}} & 99.94   \\ \cline{1-1} \cline{3-3} \cline{5-5} \cline{7-7} 
TJG-RT      & \multicolumn{1}{c|}{}                       & 99.53   & \multicolumn{1}{c|}{}                       & 99.21   & \multicolumn{1}{c|}{}                       & 99.94   \\ \cline{1-1} \cline{3-3} \cline{5-5} \cline{7-7} 
TJG-CT      & \multicolumn{1}{c|}{}                       & 99.86   & \multicolumn{1}{c|}{}                       & 100.0   & \multicolumn{1}{c|}{}                       & 100.0   \\ \hline
\end{tabular}
}
\label{table3}
\end{table}

Table \ref{table3} shows the performance comparison of each pre-trained backdoor model under different datasets, where the Clean Model is a model trained on a normal dataset. The performance of BadNet 's BA and ASR on the three datasets is relatively stable. Compared with the normal model, its BA decreases by about 1\% on the CIFAR10 and MS-Celeb-1M datasets. BadNet has the strongest data transformations with fixed visible pixel block triggers, which can maintain its attack success rate at 100\%.

WaNet uses a noise model to train the model. Compared with the normal model, its advantages can increase BA by about 1\% -1.5\%, but its disadvantages are that it requires multiple iterations and trigger data transformations is fragile. WaNet can reach more than 98\% ASR on the GTSRB and MS-Celeb-1M, but the ASR after using the ResNet18 model on the CIFAR10 is only 90.98\%.

Compared with TrojanGan and WaNet, Poison Ink has great advantages in trigger steganography performance, and trigger data transformations are enhanced. However, after the number of model training iterations is limited, compared with the normal model, its BA is reduced by about 1\% on the CIFAR10 and MS-Celeb-1M. Compared with Poison Ink's original paper, its ASR on the CIFAR-10 dataset decreased by about 0.06\%, while ASR on the MS-Celeb-1M was only 98.55\%.

The BA of the TrojanGan backdoor model ( pre-trained backdoor model ) is about 1\% lower than the normal model on the CIFAR10 and MS-Celeb-1M and is similar to other backdoor models on the GTSRB. The data overlap characteristics of the TrojanGan single trigger and the combined trigger in the same area enhance the data transformations of the trigger during the training of multi-type datasets, and offset the decline of ASR to a certain extent, making the average attack success rate of multi-target ASR more than 99.7\%. Compared with other backdoor attack models, there is no significant decline, which proves that the invisible multi-target attack of the TrojanGan backdoor model is effective.

\subsubsection{Federated Learning Backdoor Performance Analysis}
Federated learning dual model replacement attack default settings : Federated learning backdoor attack experiments were conducted on three datasets : CIFAR10, GTSRB, and MS-Celeb-1M. The local dataset owned by the participants is divided equally according to the number of participants. The training model structure uses ResNet18. The number of federal learning participants is set to 10. To reduce the training error, all participants are involved in each round of training. One participant is fixed to use dual model replacement for backdoor attack. The TrojanGan backdoor model trained in Section Pre-trained Model Performance Analysis is used as the pre-training model. The default algorithm hyperparameters $\alpha  = 0.2$,$\beta  = 6$, and the remaining participants perform normal local training. The server defaults to the average aggregation algorithm. The number of local training of federal learning is set to 4, and the number of global training is set to 35. The local learning rate of normal participants is lr = 0.01 in the first 20 rounds of global training and then adjusted to 0.001. The optimizer selects SGD, momentum = 0.9, weight\_decay = 5e-4. After each round of aggregation training, the model with the best performance on the poisoning test set is selected to save.

In the DBA experimental setup, the scaling parameter is 45, and the rest uses the default settings given in the DBA paper.

\textbf{Effectiveness analysis of DMR-FLBA algorithm}:The effectiveness analysis of the DMR-FLBA algorithm uses the federated learning dual model replacement attack default setting and uses DMR-LT, DMR-RT, and DMR-CT to represent the left trigger, right trigger, and combined trigger identifiers.

\begin{figure*}[htbp]
\centering
\subfloat{
    \includegraphics[scale=1]{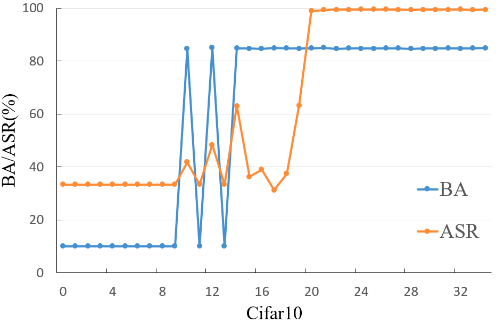}
}
\subfloat{
    \includegraphics[scale=1]{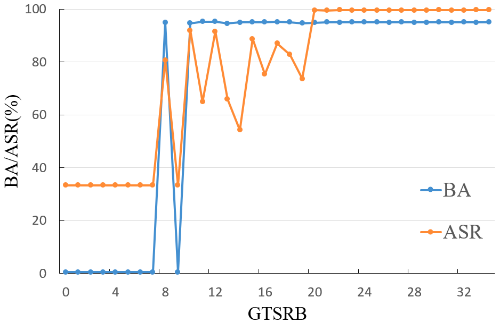}
}
\subfloat{
    \includegraphics[scale=1]{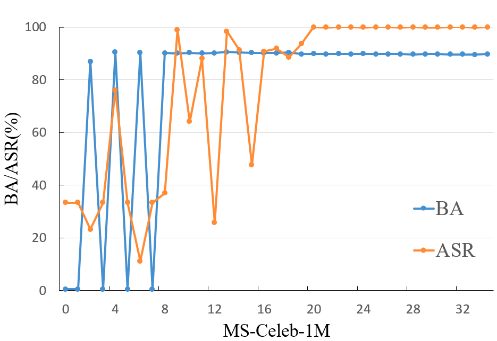}
}
\caption{\centering{DMR-FLBA attack training process}}
\label{fig5}
\end{figure*}
Fig.\ref{fig5} shows the attack training process of the DMR-FLBA aggregation model on different datasets, where BA represents the classification success rate on the normal test set, and ASR represents the average attack success rate of three different types of triggers on the poisoning test set. According to Fig.\ref{fig5} , the training of DMR-FLBA aggregation model is divided into three stages. The first stage is the latent stage (1-10 rounds). At this stage, each local model performs preliminary iterative learning. The model replacement method fails, the global aggregation parameters are chaotic, and the aggregation model cannot perform image classification. The second stage is the oscillation stage (11-21 rounds). At this stage, each local model is still not fully fitted, but the model replacement gradually takes effect, and the classification performance of the aggregation model begins to improve. The third stage is the fitting stage (21-35 rounds). The model replacement is completed, and the performance of the aggregated model is fully fitted to the pre-trained model.

Table \ref{table4} shows the performance comparison of each federated learning backdoor model under different datasets, where Clean F-Model is a federated learning model trained on normal datasets. The ASR of DBA performs poorly on the CIFAR10 and GTSRB datasets, 98.64\% and 96.85\%, respectively. After changing the trigger size, the ASR of DBA on the MS-Celeb-1M dataset can reach 99.69\%. The BA of DBA is quite different from that of each baseline model, and the average difference between the three datasets is reduced by about 5\%-10\%.
\begin{table}[ht]
\caption{\centering{Performance comparison of federated learning backdoor model}}
\centering
\scalebox{0.9}{
\begin{tabular}{|c|cc|cc|cc|}
\hline
DataSets      & \multicolumn{2}{c|}{CIFAR10}                          & \multicolumn{2}{c|}{GTSRB}                            & \multicolumn{2}{c|}{MS-Celeb-1M}                      \\ \hline
Model         & \multicolumn{1}{c|}{BA(\%)}                 & ASR(\%) & \multicolumn{1}{c|}{BA(\%)}                 & ASR(\%) & \multicolumn{1}{c|}{BA(\%)}                 & ASR(\%) \\ \hline
Clean F-Model & \multicolumn{1}{c|}{85.31}                  & -       & \multicolumn{1}{c|}{95.38}                  & -       & \multicolumn{1}{c|}{87.90}                  & -       \\ \hline
DBA           & \multicolumn{1}{c|}{69.59}                  & 98.64   & \multicolumn{1}{c|}{87.76}                  & 96.85   & \multicolumn{1}{c|}{74.90}                  & 99.69   \\ \hline
CBA           & \multicolumn{1}{c|}{80.00}                  & 100.0   & \multicolumn{1}{c|}{-}                      & -       & \multicolumn{1}{c|}{-}                      & -       \\ \hline
LR-BA         & \multicolumn{1}{c|}{79.90}                  & 98.20   & \multicolumn{1}{c|}{-}                      & -       & \multicolumn{1}{c|}{-}                      & -       \\ \hline
DMR-LT        & \multicolumn{1}{c|}{\multirow{3}{*}{84.65}} & 99.60   & \multicolumn{1}{c|}{\multirow{3}{*}{94.92}} & 99.70   & \multicolumn{1}{c|}{\multirow{3}{*}{89.75}} & 99.88   \\ \cline{1-1} \cline{3-3} \cline{5-5} \cline{7-7} 
DMR-RT        & \multicolumn{1}{c|}{}                       & 98.93   & \multicolumn{1}{c|}{}                       & 99.21   & \multicolumn{1}{c|}{}                       & 99.94   \\ \cline{1-1} \cline{3-3} \cline{5-5} \cline{7-7} 
DMR-CT        & \multicolumn{1}{c|}{}                       & 99.86   & \multicolumn{1}{c|}{}                       & 100.0   & \multicolumn{1}{c|}{}                       & 99.94   \\ \hline
\end{tabular}
}
\label{table4}
\end{table}

CBA uses distributed triggers as DBA but improves the performance of triggers. Its advantage is that the ASR of the aggregation model can reach 100\% in the CIFAR-10, but BA still has a large room for improvement. Its shortcomings are the same as DBA, that is, the number of poisoned terminals is at least 4 to achieve the expected attack performance, and the backdoor trigger is visible.

The performance of LR-BA is between DBA and CBA. Its advantage is that only a small amount of poisoning data is needed. Like DMR, the backdoor can be implanted in the aggregation model by modifying the model parameters, and the use of a 9-pixel pattern backdoor trigger has an excellent steganography effect. The disadvantage is that BA and ASR still have room for improvement.

The BA of the DMR aggregation model decreases by about 1\% on the CIFAR10 and GTSRB compared with the normal model, and increases by 2\% on the MS-Celeb-1M. The dual model replacement method makes the aggregation model retain the ASR and BA of the pre-trained model to the greatest extent, and its performance is almost unchanged compared with the pre-trained model. The average ASR of the DMR aggregation model on the CIFAR10 is 99.46\%, and the BA is 84.65\%, which is better than other baseline models, and its performance on the GTSRB and MS-Celeb-1M is also excellent. Experiments show that the DMR-FLBA backdoor attack algorithm is effective.

\begin{figure*}[htbp]
\centering
\subfloat{
    \includegraphics[scale=0.35]{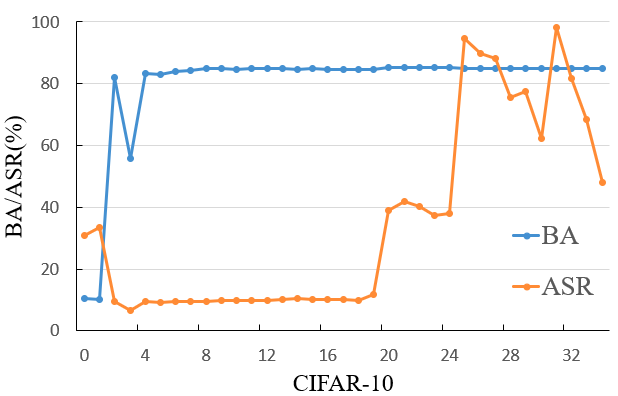}
}
\subfloat{
    \includegraphics[scale=0.35]{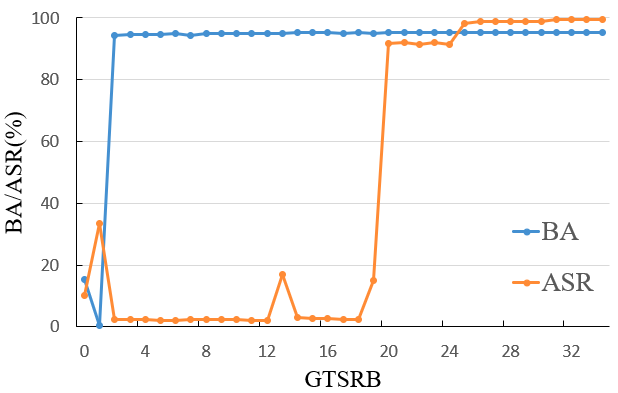}
}
\subfloat{
    \includegraphics[scale=0.35]{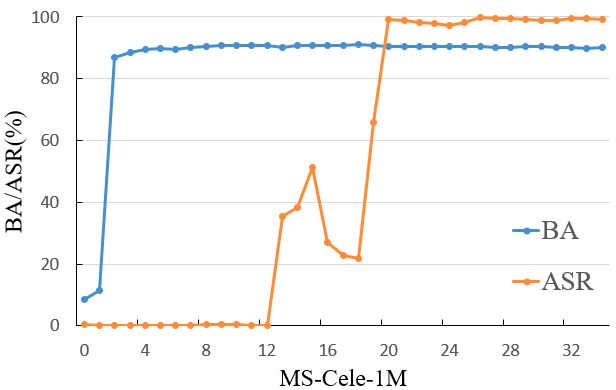}
}
\caption{\centering{Attack training process under the influence of attack interval}}
\label{fig6}
\end{figure*}

\begin{figure*}[htbp]
\centering
\subfloat{
    \includegraphics[scale=0.35]{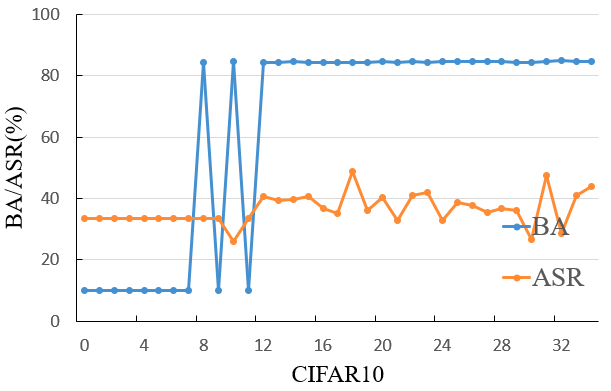}
}
\subfloat{
    \includegraphics[scale=0.35]{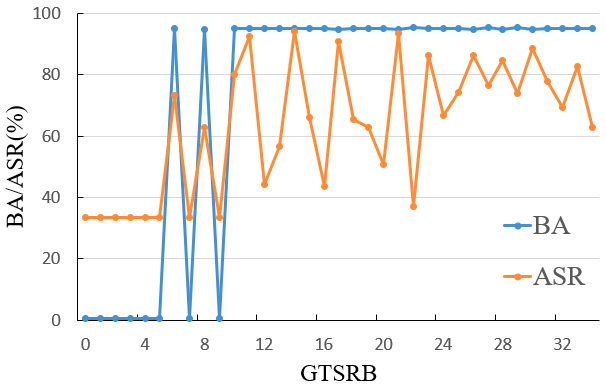}
}
\subfloat{
    \includegraphics[scale=0.35]{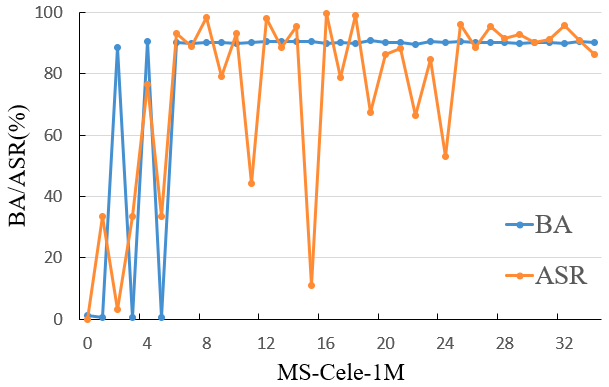}
}
\caption{\centering{Attack training process under the influence of learning rate}}
\label{fig7}
\end{figure*}

\textbf{Influence analysis of $\alpha $ in DMR-FLBA algorithm}:The influence experiment of hyperparameter $\alpha $ gradually adjusts the influence value of $\alpha$ based on the default setting of federated learning dual model replacement attack. Like the following, the ASR in Table \ref{table5} represents the average attack success rate of three different backdoors. Table \ref{table5}  shows the changes in the performance of the aggregation model with the change of the $\alpha $ under three different datasets. In general, with the increase of the  $\alpha $, the proportion of the aggregation model parameters will be integrated into the local backdoor model will increase, and the BA of the aggregation model will continue to increase, resulting in a negative effect that the ASR will continue to decline. When the $\alpha $ increase to more than 1, BA will become relatively stable, but ASR will decrease to 10 \%, and the backdoor of the aggregation model will completely fail. 

\begin{table}[ht]
\caption{\centering{The influence of $\alpha $ changes}}
\centering
\scalebox{0.9}{
\begin{tabular}{|c|cc|cc|cc|}
\hline
DataSets  & \multicolumn{2}{c|}{CIFAR10}          & \multicolumn{2}{c|}{GTSRB}            & \multicolumn{2}{c|}{MS-Celeb-1M}      \\ \hline
Parameter & \multicolumn{1}{c|}{BA(\%)} & ASR(\%) & \multicolumn{1}{c|}{BA(\%)} & ASR(\%) & \multicolumn{1}{c|}{BA(\%)} & ASR(\%) \\ \hline
$\alpha  = 0$         & \multicolumn{1}{c|}{84.61}  & 99.68   & \multicolumn{1}{c|}{94.91}  & 99.67   & \multicolumn{1}{c|}{89.70}  & 99.92   \\ \hline
$\alpha  = 0.2$          & \multicolumn{1}{c|}{84.65}  & 99.46   & \multicolumn{1}{c|}{94.92}  & 99.64   & \multicolumn{1}{c|}{89.75}  & 99.92   \\ \hline
$\alpha  = 0.4$         & \multicolumn{1}{c|}{84.83}  & 99.08   & \multicolumn{1}{c|}{94.93}  & 99.57   & \multicolumn{1}{c|}{89.80}  & 99.92   \\ \hline
$\alpha  = 0.6$          & \multicolumn{1}{c|}{84.75}  & 99.02   & \multicolumn{1}{c|}{94.92}  & 99.57   & \multicolumn{1}{c|}{89.90}  & 99.88   \\ \hline
$\alpha  = 0.8$          & \multicolumn{1}{c|}{84.95}  & 97.57   & \multicolumn{1}{c|}{95.11}  & 99.44   & \multicolumn{1}{c|}{89.85}  & 99.62   \\ \hline
\end{tabular}
}
\label{table5}
\end{table}

\textbf{Influence analysis of $\beta$ in DMR-FLBA algorithm}:The influence experiment of hyperparameter $\beta$ gradually adjusts the influence value of $\beta$ based on the default setting of federated learning dual model replacement attack. Table \ref{table6} shows the change in the performance of the aggregation model with the change of $\beta$ under three different datasets. In general, with the increase of $\beta$, the parameter values submitted by the local backdoor model to the aggregation model increase exponentially. The effect of $\beta$ is opposite to that of $\alpha $. The ASR of the aggregation model will continue to increase, resulting in a negative effect that BA will continue to decline. When $\beta$ increases to more than the number of participants in federal learning, the dual model replacement will cause negative effects. At this time, the backdoor cannot be implanted into the aggregation model, and the aggregation model BA will remain at 10\%.

\begin{table}[ht]
\caption{\centering{The influence of $\beta$ changes}}
\centering
\scalebox{0.9}{
\begin{tabular}{|c|ll|ll|ll|}
\hline
DataSets  & \multicolumn{2}{c|}{CIFAR10}                               & \multicolumn{2}{c|}{GTSRB}                                 & \multicolumn{2}{c|}{MS-Celeb-1M}                           \\ \hline
Parameter & \multicolumn{1}{c|}{BA(\%)} & \multicolumn{1}{c|}{ASR(\%)} & \multicolumn{1}{c|}{BA(\%)} & \multicolumn{1}{c|}{ASR(\%)} & \multicolumn{1}{c|}{BA(\%)} & \multicolumn{1}{c|}{ASR(\%)} \\ \hline
$\beta  = 1$         & \multicolumn{1}{l|}{84.96}  & 94.02                        & \multicolumn{1}{l|}{95.20}  & 98.56                        & \multicolumn{1}{l|}{90.14}  & 99.42                        \\ \hline
$\beta  = 2$         & \multicolumn{1}{l|}{84.87}  & 98.71                        & \multicolumn{1}{l|}{95.02}  & 99.51                        & \multicolumn{1}{l|}{89.95}  & 99.66                        \\ \hline
$\beta  = 4$         & \multicolumn{1}{l|}{84.86}  & 99.20                        & \multicolumn{1}{l|}{94.93}  & 99.54                        & \multicolumn{1}{l|}{89.75}  & 99.92                        \\ \hline
$\beta  = 6$         & \multicolumn{1}{l|}{84.65}  & 99.46                        & \multicolumn{1}{l|}{94.92}  & 99.64                        & \multicolumn{1}{l|}{89.75}  & 99.92                        \\ \hline
$\beta  = 8$         & \multicolumn{1}{l|}{84.45}  & 99.66                        & \multicolumn{1}{l|}{94.88}  & 99.67                        & \multicolumn{1}{l|}{89.50}  & 99.98                        \\ \hline
\end{tabular}
}
\label{table6}
\end{table}

\textbf{Influence analysis of attack participants' number}:The influence experiment of the number of attack participants is based on the federal learning dual model replacement attack default setting. To observe the performance change of the aggregation model more easily, the hyperparameter $\beta  = 1$ is reset and the number of attack participants is gradually adjusted. Table \ref{table7} shows the increase in the number of attack participants and the change in the performance of the aggregation model under three different datasets. In general, the increase in the number of attack participants, the increase in the proportion of parameters submitted to the aggregation model by multiple local backdoor models, the ASR of the aggregation model will continue to increase, and the negative effect of the decrease in normal participants is that the BA will continue to decline. However, even in the case of hyperparameter $\beta  = 1$ and only one attack participant, the average ASR of the three datasets reaches 97.33\%, which proves that the attack of the first model replacement is very effective.

\begin{table}[ht]
\caption{\centering{The influence of attack participants' number}}
\centering
\scalebox{0.9}{
\begin{tabular}{|c|ll|ll|ll|}
\hline
DataSets     & \multicolumn{2}{c|}{CIFAR10}                               & \multicolumn{2}{c|}{GTSRB}                                 & \multicolumn{2}{c|}{MS-Celeb-1M}                           \\ \hline
Participants & \multicolumn{1}{c|}{BA(\%)} & \multicolumn{1}{c|}{ASR(\%)} & \multicolumn{1}{c|}{BA(\%)} & \multicolumn{1}{c|}{ASR(\%)} & \multicolumn{1}{c|}{BA(\%)} & \multicolumn{1}{c|}{ASR(\%)} \\ \hline
1            & \multicolumn{1}{l|}{84.96}  & 94.02                        & \multicolumn{1}{l|}{95.20}  & 98.56                        & \multicolumn{1}{l|}{90.14}  & 99.42                        \\ \hline
2            & \multicolumn{1}{l|}{84.94}  & 98.66                        & \multicolumn{1}{l|}{95.04}  & 99.44                        & \multicolumn{1}{l|}{89.90}  & 99.83                        \\ \hline
3            & \multicolumn{1}{l|}{84.80}  & 99.13                        & \multicolumn{1}{l|}{95.03}  & 99.57                        & \multicolumn{1}{l|}{89.80}  & 99.92                        \\ \hline
4            & \multicolumn{1}{l|}{84.77}  & 99.24                        & \multicolumn{1}{l|}{94.93}  & 99.64                        & \multicolumn{1}{l|}{89.85}  & 99.92                        \\ \hline
\end{tabular}
}
\label{table7}
\end{table}

\textbf{Influence analysis of attack interval}:The influence experiment of the number of attack participants adjusts the attack interval based on the federal learning dual model replacement attack default setting. Table \ref{table8} shows the change in the performance of the aggregation model with the increase of the attack interval under three different datasets. In general, the attack interval increases, the frequency of the local backdoor model submitting to the backdoor parameter of the aggregation model decreases, the ASR of the aggregation model will continue to decrease, and the effect of the increase of normal participants is that the BA will increase. Fig.\ref{fig6} shows the training process of the dual model replacement attack when the attack interval is 5. It can be seen from Fig.\ref{fig6} that the dual model replacement has a very limited impact on the first 20 rounds of model training when the attack interval is large. After 20 rounds, the backdoor begins to be successfully implanted into the aggregation model, and the ASR of the aggregation model rises rapidly. However, if there is no poisoning parameter uploaded within a certain interval, the ASR will show a gradual downward trend, which is particularly evident on the CIFAR10.

\begin{table}[ht]
\caption{\centering{Impact of attack interval change}}
\centering
\scalebox{0.9}{
\begin{tabular}{|c|ll|ll|ll|}
\hline
DataSets & \multicolumn{2}{c|}{CIFAR10}                               & \multicolumn{2}{c|}{GTSRB}                                 & \multicolumn{2}{c|}{MS-Celeb-1M}                           \\ \hline
Interval & \multicolumn{1}{c|}{BA(\%)} & \multicolumn{1}{c|}{ASR(\%)} & \multicolumn{1}{c|}{BA(\%)} & \multicolumn{1}{c|}{ASR(\%)} & \multicolumn{1}{c|}{BA(\%)} & \multicolumn{1}{c|}{ASR(\%)} \\ \hline
0        & \multicolumn{1}{l|}{84.65}  & 99.46                        & \multicolumn{1}{l|}{94.92}  & 99.64                        & \multicolumn{1}{l|}{89.75}  & 99.92                        \\ \hline
1        & \multicolumn{1}{l|}{84.76}  & 99.26                        & \multicolumn{1}{l|}{95.04}  & 99.67                        & \multicolumn{1}{l|}{89.40}  & 99.85                        \\ \hline
3        & \multicolumn{1}{l|}{84.81}  & 98.91                        & \multicolumn{1}{l|}{95.02}  & 99.61                        & \multicolumn{1}{l|}{89.64}  & 99.86                        \\ \hline
5        & \multicolumn{1}{l|}{85.00}  & 98.31                        & \multicolumn{1}{l|}{95.11}  & 99.50                        & \multicolumn{1}{l|}{89.90}  & 99.75                        \\ \hline
\end{tabular}
}
\label{table8}
\end{table}

\textbf{Influence analysis of local learning rate}:The local learning rate influence experiment is based on the default setting of the federal learning dual model replacement attack. The local learning rate hyperparameter lr = 0.01 is fixed, and the learning rate is not adjusted halfway. Fig.\ref{fig7} shows the changes in the performance of the aggregation model with a fixed local learning rate under three different datasets. In general, the BA of the aggregation model decreases by about 1\% -5\%, the ASR decreases to varying degrees and the training process shows a trend of oscillation. For the CIFAR-10, its ASR decreases by about 60\%, and for the GTSRB and MS-Celeb-1M, its ASR decreases by about 1\%. The reason for this situation is that local training cannot learn the expected parameters because of the fixed learning rate, which leads to the failure of the aggregated model after the pre-training model replacement to fit and ultimately leads to the failure of backdoor injection. This situation is particularly serious for CIFAR10 which requires lower learning rate training. Here, a disadvantage of the dual model replacement method is missed, that is, if the federated learning local model fails to reach the minimum learning rate during the pre-trained model training, the model replacement may fail in the backdoor injection. 

\section{Conclusion}
This paper comprehensively considers the three stages of trigger generation, data poisoning, and model training in backdoor attack. Starting from the three aspects of trigger steganography, multi-target attack, and model poisoning, according to the characteristics of federated learning training, an invisible multi-target backdoor attack method under federated learning is designed. Firstly, this paper designs a TrojanGan model to generate a steganography trigger, which balances the steganography performance and recognition performance of the trigger. Secondly, a multi-type poisoning dataset is created for multi-target attacks and a pre-training model is trained. Finally, a dual model replacement method is designed to improve the attack and recognition performance of the federated learning aggregation model. The experimental results show that the DMR-FLBA algorithm can improve the accuracy of identifying normal data while ensuring the high backdoor attack success rate of the aggregation model. In future work, we will explore backdoor defense methods for dual model replacement method in federated learning.

\textbf{Acknowledgement.}This work was supported in part by the Youth Innovation Group Support Program of ICE Discipline of CQUPT (No.SCIE-QN-2022-05)


%





\ifCLASSOPTIONcaptionsoff
  \newpage
\fi

\end{document}